AGENT-O: A Semantic Agent Card Framework for Interoperable and Governed Healthcare AI Agents

**Authors and institutions**

Pengze Li, PhD[1], Cui Tao, PhD[1,#]

**Author Affiliations:**

1-Department of Artificial Intelligence and Informatics, Mayo Clinic, Jacksonville, FL, USA

# **Correspondence**:

Cui Tao, PhD **tao.cui@mayo.edu**

Address: 4500 San Pablo Rd S., Jacksonville, FL 32224 USA

## ABSTRACT

### Objective

To develop and evaluate AGENT-O, a modular ontology framework that defines a semantic Agent Card for representing health-oriented AI agent systems and supports assessment of agent-system reporting completeness in scientific publications.

### Materials and Methods

AGENT-O was developed as a modular OWL 2/RDF ontology covering runtime, models, workflow, tools, clinical use, evaluation, provenance, governance, and reporting assessment. Evaluation included ontology inventory, OWL-RL reasoning, three SHACL suites, 12 SPARQL queries, three cases, and model-assisted reporting-completeness assessment of 279 papers across five dimensions.

### Results

The ontology contained 1,962 RDF triples and 1,922 Protégé axioms, with 252 active classes, 198 active object properties, and 51 datatype properties. All SHACL suites conformed on example graphs, all competency queries returned prespecified evidence, and all 279 papers were scored. Incomplete reporting was highest for runtime/architecture (84.6%), governance/safety (82.8%), and provenance/reproducibility (78.1%), versus evaluation (25.8%) and benchmark-process alignment (29.8%).

### Discussion

AGENT-O supported semantic Agent Card representation and reporting assessment while revealing an evaluation–specification gap: evaluation and benchmark procedures were reported more consistently than runtime architecture, governance, and reproducibility.

### Conclusion

AGENT-O provides a reusable ontology, semantic Agent Card profile, and reporting-completeness workflow. It supports structured reporting and gap identification but does not assess agent quality or deployment readiness.

## Background and Significance

With rapid advances in artificial intelligence (AI), healthcare AI encompasses a broad range of predictive, generative, multimodal, conversational, decision-support, and workflow-automation systems.[1] Within this broader landscape, an emerging class of systems is better understood as health-oriented AI agents: systems that use one or more AI models to pursue health-related goals through iterative planning, observation, tool use, workflow orchestration, and interaction with human users or clinical environments.[2-4]

Although the concept of an agent is relatively well established in artificial intelligence, the rapid adoption of agentic AI in healthcare has introduced substantial variation in what constitutes an "AI agent" and how such systems are implemented and evaluated.[5,6] Existing reporting frameworks for non-agentic health AI generally focus on data sources, model development, performance evaluation, and clinical evaluation.[7,8] These elements remain necessary for agentic AI but do not fully characterize how an agent functions in practice. Agentic systems may observe and respond to dynamic information, access external tools and resources, take actions within clinical or research workflows, and interact with humans or other agents.[5,6] Consequently, transparent reporting requires additional information about an agent's operational boundaries, including what it can observe, which tools and resources it can access, what actions it is authorized to take, and where human review, override, or escalation occurs. In healthcare, these details are essential for assessing reproducibility, safety, accountability, and the conditions under which reported findings may generalize to other settings. [8,9]

Existing health AI reporting guidelines have improved transparency for clinical trial protocols, early-stage evaluations, and prediction-model studies.[7-11] In parallel, documentation frameworks such as model cards and datasheets have promoted structured reporting of model and dataset characteristics, including intended use, performance, limitations, data provenance, and ethical considerations.[12-14] Recent domain-specific efforts such as ROADMAP extend model-card and datasheet concepts with machine-interpretable descriptions of medical AI models and datasets.[15] Croissant similarly standardizes machine-readable metadata for ML datasets, including their contents, provenance, and usage restrictions.[16] These resources serve complementary purposes, but none was designed to represent an AI agent as an integrated, tool-mediated system with runtime actions, intermediate artifacts, human interactions, and governance constraints. Consequently, published studies may report model performance while leaving key aspects of agent design, operation, provenance, and governance insufficiently specified. [5,6,12-14]

Ontologies provide a promising approach to addressing this gap.[17-19] Semantic Web standards, including RDF[20], OWL[21], SPARQL[22,23], and SHACL[24] enable formal representation, querying, and validation of structured knowledge. Existing biomedical and governance resources address important adjacent domains, including provenance[25], clinical data representation[26], software description[27], bibliographic and scientific evidence[28,29], model reporting[7,10,12], and AI risk management[30,31]. However, the concepts needed for health-oriented AI agent reporting remain distributed across these resources. No single resource provides a unified framework for representing an agent's architecture, execution process, paper-based evidence, and governance-relevant information.

We introduce AGENT-O, an ontology-based framework for representing health-oriented AI agent systems, including components, workflows, execution context, model resources and deployments, clinical use, evaluation, provenance, and governance. Its semantic Agent Card is an ontology-conformant system-reporting profile, and its reporting-completeness workflow identifies whether publications provide information needed for reproducible, queryable annotation. AGENT-O is intended as a reusable framework rather than a universal ontology. We evaluated whether AGENT-O supports structured system representation, identifies publication-level reporting gaps, and connects semantic interoperability with reporting assessment.

# Materials and Methods

## Scope and Design Principles

AGENT-O was developed as a modular OWL 2 ontology to support structured representation and reporting assessment of health-oriented AI agent systems.[21] The ontology was designed to capture both the characteristics of agent systems and the reporting elements needed to describe those systems in a structured and consistent manner. Its scope includes agent identity, runtime architecture, model components and their functional roles, versioned AI model specifications, runtime model deployments, workflow, planning and reasoning, memory, tool use, execution context, generated outputs and proposed actions, evaluation design, provenance, governance, clinical context and intended use, health-data interoperability, reporting-assessment provenance and evidence, and an ontology-conformant Agent Card profile that assembles these elements into a system-level report.

Within AGENT-O, an Agent Card is an ontology-conformant report describing an AgentSystem. It organizes system identity, model roles, specifications and deployments, workflow and tools, clinical use, evaluation, provenance, governance, limitations, and evidence. The report remains distinct from the system it describes. AGENT-O also separates model role from specification and deployment, clinical use from manuscript statements, and assessment records from the external workflow that creates them.

Development combined conceptual modeling, alignment with established resources, competency queries, SHACL profile validation, and reporting-completeness assessment. Four principles guided the work: represent systems at reproducibility-oriented detail; reuse rather than redefine established concepts; separate system entities from publication descriptions and assessment activities; and make governance, provenance, evidence, and reporting gaps queryable.

Figure 1 summarizes source review, ontology construction and alignment, semantic validation, competency-query testing, case and corpus assessment, and feedback-driven refinement.

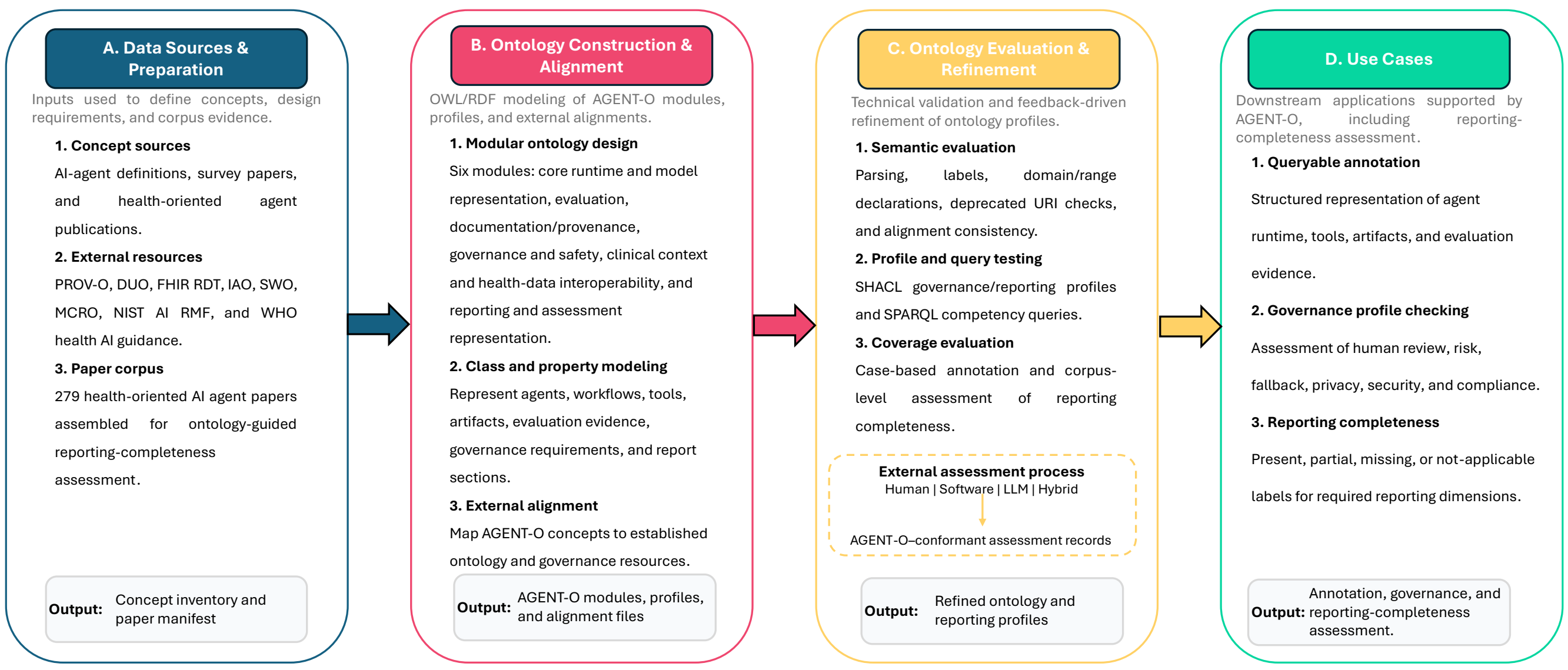

Figure 1. Overview of the AGENT-O development pipeline: data sources and preparation, ontology construction and external alignment, ontology evaluation and refinement, and use cases.

## Data sources

AGENT-O was informed by three categories of source material. First, we reviewed definitions and conceptual descriptions of AI agents from survey and review papers to establish an initial concept inventory.[2-4] These sources informed baseline concepts related to agent identity, workflow, planning, reasoning, memory, tool use, execution, proposed actions, evaluation, governance, provenance, lineage, uncertainty, risk, security, privacy, and compliance.

Second, we reviewed biomedical, provenance, software, reporting, and governance ontologies or standards to identify concepts suitable for reuse or alignment. PROV-O was used for provenance representation.[25] The Data Use Ontology addressed data-use permissions, restrictions, and conditions.[32] FHIR RDF provided representations for health-data resources such as Patient, Observation, Condition, Procedure, DiagnosticReport, Medication, and MedicationRequest.[26] The Information Artifact Ontology supported the representation of information artifacts and document sections.[33] The Software Ontology covered software, versions, licenses, and implementation-related entities.[27] The Model Card Report Ontology contributed model-card and model-reporting concepts.[13] The NIST AI Risk Management Framework informed governance, mapping, measurement, and management functions.[31] WHO guidance on health AI informed principles related to autonomy, transparency, accountability, equity, safety, and privacy[30,34]. STATO was additionally reviewed for its formal representation of statistical methods, probability distributions, experimental design, and result-reporting concepts[35]. The HL7 AI Transparency on FHIR Implementation Guide was reviewed as a complementary implementation-level resource for representing AI-influenced data, model documentation, provenance, input data sources, and human–AI processes in FHIR[36]. These resources were treated as complementary standards with which AGENT-O should interoperate, rather than as competing ontologies.

Third, we assembled a curated-source corpus from the literature inventories compiled for a systematic review and taxonomy of agentic large-language-model systems in medicine and a comprehensive survey of AI agents in healthcare.[37,38] The final corpus comprised 278 extracted paper documents plus the prespecified AgentArena case, for 279 records. Because this review-derived corpus was not based on an independent systematic bibliographic-database search, it was not intended to exhaustively cover the health-oriented AI-agent literature.

## Ontology design

### Conceptual model and class architecture

AGENT-O represents agent systems through six linked domains: core runtime and model representation, evaluation, documentation and provenance, governance and safety, clinical context and health-data interoperability, and reporting and assessment. Three boundaries prevent common conflations: model role versus specification versus deployment; model intended use versus system clinical use versus manuscript statements; and ontology records of an assessment versus the external workflow that executes it (Figure 2).

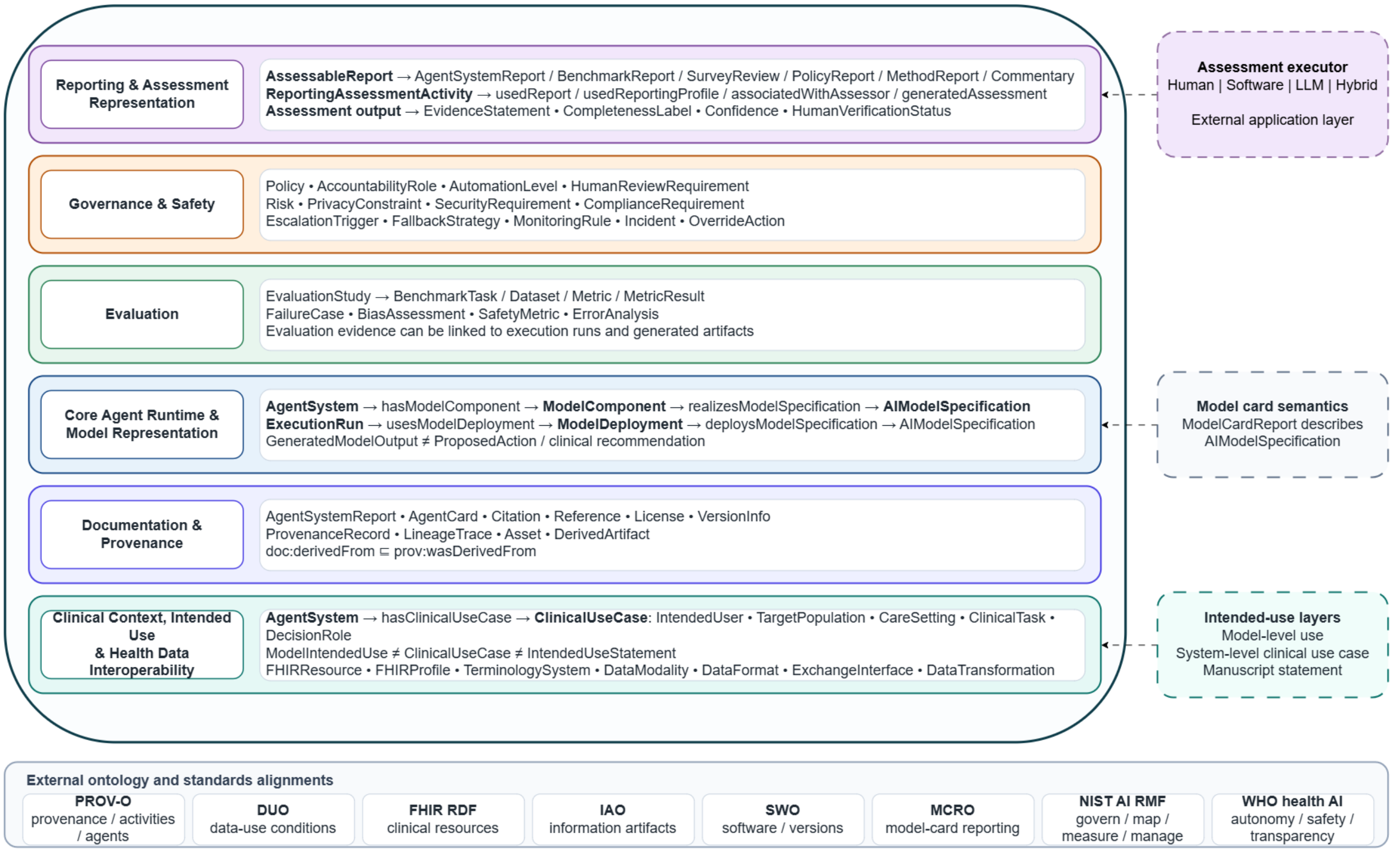

**Figure 2. AGENT-O modular class architecture and major semantic relationships.** Representative classes and relations are shown across six modules: core runtime and model representation, evaluation, documentation and provenance, governance and safety, clinical context and health-data interoperability, and reporting and assessment representation. Solid connectors denote native AGENT-O relations. Dashed connectors denote cross-boundary links, while dashed callout boxes identify either the external assessment layer or explanatory semantic distinctions. External ontology and standards mappings are summarized in the bottom alignment band. Selected entities are shown for clarity rather than the complete OWL class hierarchy.

The core module separates ModelComponent (a functional role), AIModelSpecification (a versioned model resource), and ModelDeployment (the endpoint or checkpoint used during execution). Specifications may record identity, architecture, developer, interfaces, modalities, capabilities, intended use, and limitations; deployments may record provider, version, access time, serving configuration, region, and runtime settings.

A ModelComponent realizes a specification, whereas an ExecutionRun uses a deployment that deploys a specification. Generated outputs are distinct from proposed actions, and the three model classes are disjoint.

The evaluation module links studies to tasks, datasets, metrics, results, failures, bias or safety assessments, and execution evidence.

The documentation and provenance module represents references, lineage, assets, licenses, versions, code, data, models, and derived artifacts, with alignment to PROV-O, IAO, and SWO.

The governance and safety module makes policies, accountability, automation, human review, risk, privacy, security, compliance, escalation, fallback, monitoring, incidents, and overrides queryable.

The clinical-context module represents ClinicalUseCase entities with users, populations, settings, tasks, conditions, workflow stages, intended actions, decision roles, deployment modes, and excluded uses.

It also separates model-level intended use, system-level clinical use, and manuscript statements, while representing FHIR-aligned resources, terminology, modalities, interfaces, and transformations.

The reporting module distinguishes AgentSystemReport from benchmark, review, policy, method, and commentary reports. Only AgentSystemReport directly documents an AgentSystem.

It represents profiles, requirements, assessments, assessors, evidence, source locations, confidence, workflow versions, completeness labels, and human-verification status. External human or software workflows perform extraction and scoring.

External alignments make AGENT-O an integrating layer rather than a replacement for established provenance, software, clinical-data, documentation, model-reporting, or governance resources.

## Relation model and property constraints

Object properties connect entities across modules; datatype properties record identifiers, versions, timestamps, scores, thresholds, confidence, and descriptions.

Runtime properties link systems, workflow steps, tools, executions, observations, outputs, and actions. Model properties connect component roles to specifications and execution runs to deployments, allowing role, identity, version, and execution configuration to be queried together.

Evaluation properties link studies, tasks, datasets, metrics, results, failures, and runs; provenance properties link artifacts to lineage, sources, licenses, versions, and PROV-O usage or generation patterns.

Clinical and governance properties connect systems to use contexts, data permissions, accountability, human review, risks, privacy, security, escalation, fallback, monitoring, incidents, and overrides.

Reporting properties link reports to sections, profiles, evidence, assessors, activities, completeness results, confidence, versions, timestamps, and verification status, keeping documentation assessment distinct from the system itself.

Domains, ranges, and disjointness constraints preserved boundaries among systems, models, deployments, and assessment activities.

### External alignment

AGENT-O alignment files map to PROV-O, DUO, FHIR RDF, IAO, SWO, MCRO, NIST AI RMF, and WHO health AI guidance.

Mappings use subclass, subproperty, exact-match, close-match, or related-match relations according to semantic correspondence.

These resources supply established semantics for provenance, data-use permissions, clinical resources, information artifacts, software, risk management, accountability, oversight, privacy, and safety.

MCRO is used specifically for model-reporting semantics: a ModelCardReport describes an AIModelSpecification but is distinct from the specification, its ModelComponent role, and its runtime ModelDeployment. AGENT-O connects these entities to agent workflows, clinical use, evaluation, governance, and reporting assessment.

## Evaluation Methods

Evaluation comprised four complementary procedures: formal content and structural checks; OWL-RL reasoning and SHACL validation; SPARQL competency queries; and case- and corpus-level reporting-completeness assessment. The first three assessed technical integrity and inferential behavior; coverage evaluation tested representability and publication completeness.

## Formal ontology content and semantic validation

All modules, profiles, alignments, shapes, and example graphs were parsed independently. Checks covered deprecated namespaces, labels, property domains and ranges, undefined alignment terms, and consistency between the integrated ontology and component modules.

RDF serialization size and Protégé axiom counts were reported separately. The integrated graph contained 1,962 triples; Protégé reported 1,922 total axioms, including 687 logical and 506 declaration axioms. Excluding deprecated entities yielded 252 active classes, 198 active object properties, and 51 active datatype properties. Complete axiom and module inventories appear in Supplementary Table S1.

OWL-RL reasoning materialized subclass, subproperty, domain, range, equivalence, inverse-property, and provenance inferences before competency-query execution.

## SHACL application-profile validation

Three SHACL suites tested architecture, governance, and reporting profiles, including model-layer distinctions, oversight and risk controls, report structure, evidence, assessor provenance, and completeness records.

Each suite was applied to its corresponding example graph; conformance therefore tested those instances against profile constraints, not global ontology completeness.

## Competency-question evaluation

Twelve SPARQL queries tested model roles, specifications and deployments; clinical use and FHIR inputs; data-use and governance; model-card and publication evidence; provenance; assessment activities; incomplete sections; and report-type scope.

A query passed when OWL-RL-materialized bindings matched prespecified entities and relations. Query text, expected bindings, row counts, evidence, and feasible negative controls were retained (Supplementary Table S4B).

## Coverage evaluation

Coverage evaluation used publication descriptions as the primary evidence source. Case-based evaluation included AgentArena as a benchmark-alignment case represented as a BenchmarkReport[39], MedAgent-Pro as an evidence-based medical-agent case[40], and a multi-agent medical decision consensus matrix system as a governance- and oversight-intensive case.[41] Each case was assessed across runtime and architecture, evaluation, provenance and reproducibility, governance and safety, and benchmark-process alignment. The purpose was to determine whether the reported agent system or benchmark framework could be represented using AGENT-O and whether the publication provided enough information to instantiate the relevant ontology elements.

Corpus-level assessment used a curated-source collection of 279 health-oriented AI agent papers assembled from the literature inventories associated with the two surveys[37,38]. AgentArena was retained as a prespecified benchmark-alignment case. Papers were categorized as agent-system, benchmark, survey/review, governance/policy, method/model, or conceptual/commentary reports. Because the corpus was derived from these two surveys rather than an independent bibliographic-database search, it was not intended to be exhaustive.

A label-blinded procedure supplied information extracted from each paper to GPT-5.1 through an institution-hosted Azure OpenAI-compatible gateway. The judge assigned report types and dimension labels and returned evidence, source locations, gaps, and confidence. Deterministic postprocessing verified evidence, corrected spans, enforced report-type applicability, and did not increase scores. Present, partial, and missing received 1, 0.5, and 0; not applicable was excluded. Rubric, workflow settings, metadata, and remaining provenance limitations are reported in the Supplementary Methods.

$$C_i = 100 \times \frac{\sum_{d \in A_i} w_d \, s_{id}}{\sum_{d \in A_i} w_d}$$

where $A_i$ is the set of dimensions applicable to paper $i$, $w_d$ is the dimension weight, and $s_{id}$ is 1.0 for present, 0.5 for partial, and 0 for missing. The weights were 25% for runtime and architecture, 25% for evaluation, 20% for provenance and reproducibility, 20% for governance and safety, and 10% for benchmark-process alignment.

Corpus-level results were summarized by paper, dimension, score band, and paper type. Reporting-completeness scores were interpreted exclusively as measures of publication-level reporting. They were not interpreted as measures of agent performance, scientific correctness, clinical utility, fairness, safety, or deployment readiness.

**Table 1. Reporting-completeness scoring dimensions.**

| Dimension | Weight | Main reporting elements assessed |
| --- | --- | --- |
| Runtime and architecture | 25% | Agent identity and scope, components and roles, workflow/planning/reasoning, memory and context, tools and execution environment, model identity/interfaces/capabilities/deployment, and clinical intended use and boundaries |
| Evaluation | 25% | Evaluation datasets and tasks, baselines and comparators, metrics and results, uncertainty and variability, and error and failure analysis |
| Provenance and reproducibility | 20% | Source-data provenance, execution traces and lineage, code/data/artifacts, versions and configurations, environments and dependencies, and access/licensing/reproduction instructions |
| Governance and safety | 20% | Governance policies and accountability, human review and override, escalation and fallback, uncertainty/refusal/abstention, risk and safety, and privacy/security/compliance/ethics/bias/fairness |
| Benchmark-process alignment | 10% | Separation of runtime and evaluation, stepwise and intermediate verification, validity and refusal handling, reliability and stability, cost and resource reporting, and task taxonomy and extensibility |

# Results

## Ontology content and semantic validation

The evaluated release contained 1,962 RDF triples. Protégé reported 1,922 total axioms, including 687 logical and 506 declaration axioms; the active inventory comprised 252 classes, 198 object properties, and 51 datatype properties across six modules. All 25 Turtle files parsed successfully, and the release included 183 external mappings and three application profiles for architecture, governance, and reporting (Supplementary Tables S1 and S2).

Figure 3 shows representative classes across six domains: (A) AgentSystem, model roles, specifications, interfaces, intended use, and deployments; (B) artifacts, execution, memory, reasoning, tools, and workflow; (C) evaluation, metrics, failures, provenance, lineage, versions, and licenses; (D) clinical use, users, populations, settings, tasks, conditions, FHIR inputs, and health-data sources; (E) automation, accountability, review, escalation, fallback, monitoring, policy, and risk; and (F) reports, assessors, evidence, assessments, completeness results, profiles, requirements, and verification status.

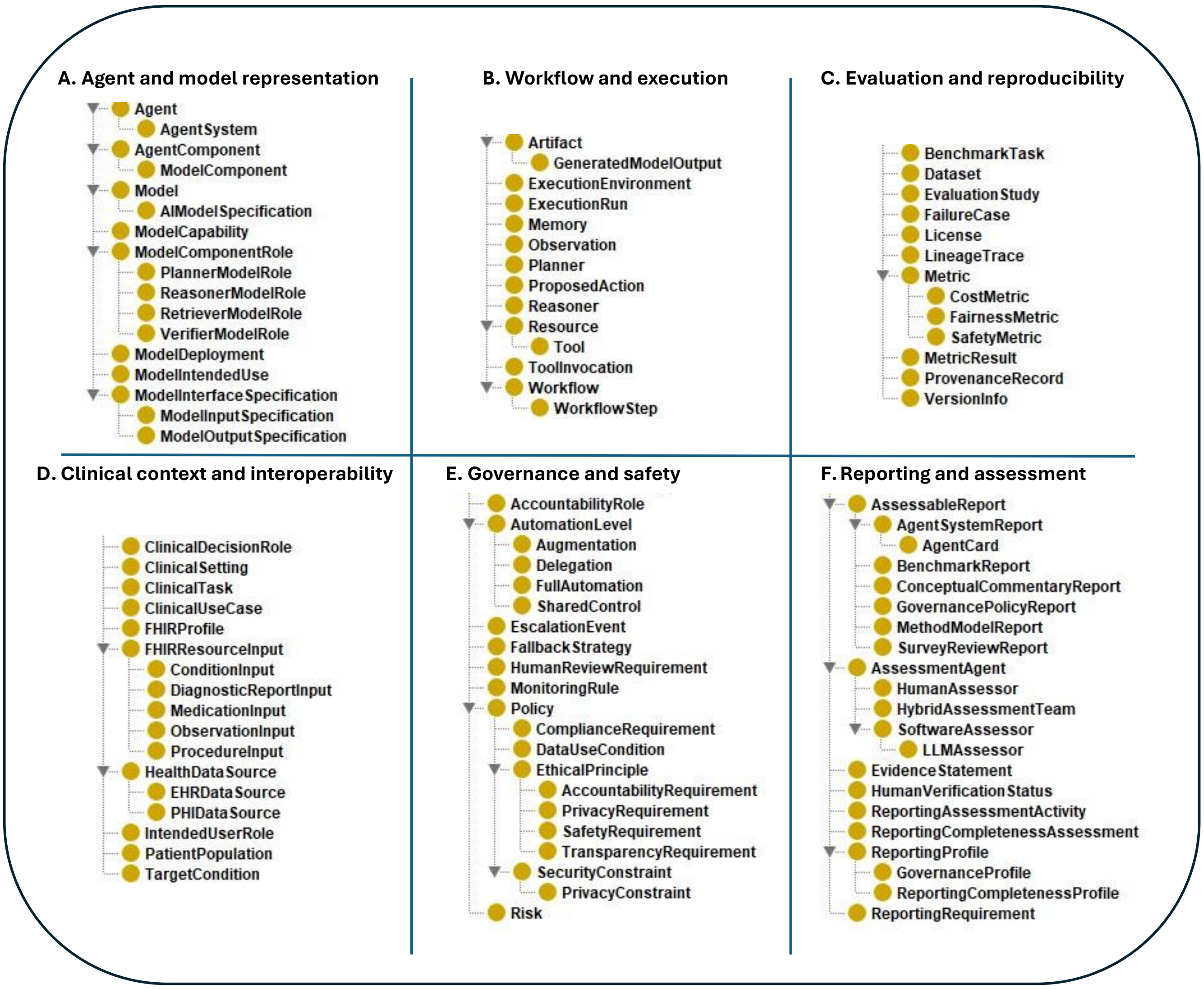

Figure 3. Representative AGENT-O class architecture across six semantic domains: **(A)** agent and model representation, **(B)** workflow and execution, **(C)** evaluation and reproducibility, **(D)** clinical context and health-data interoperability, **(E)** governance and safety, and **(F)** reporting and assessment. Selected classes are shown to illustrate major entity boundaries rather than the complete OWL hierarchy.

Automated validation found no parse failures, deprecated predecessor namespaces, missing labels or active-property domains and ranges, or undefined AGENT-O terms in alignment files (Supplementary Table S3).

All architecture, governance, and reporting SHACL suites conformed without findings on their corresponding example graphs. All 12 competency queries returned prespecified evidence for model structure, clinical use, FHIR inputs, governance, model-card alignment, provenance, and assessment records (Supplementary Tables S4A and S4B).

## Case-based reporting-completeness

Case-based reporting-completeness demonstrated that AGENT-O could distinguish ontology representability from paper-description completeness. Representative cases covered benchmark alignment[39], evidence-based medical diagnosis[40], and multi-agent medical decision support.[41] These cases were selected to test whether AGENT-O could represent runtime architecture, evaluation design, provenance and reproducibility artifacts, governance and human-review requirements, and benchmark-process alignment across different agent-paper types.

All three cases were representable while differing in reporting completeness. AgentArena, modeled as a BenchmarkReport, was partial for provenance/reproducibility and governance/safety. MedAgent-Pro reported runtime, evaluation, provenance, and benchmark procedures more completely but was partial for governance/safety. The multi-agent consensus case satisfied all five dimensions in the curated annotation (Supplementary Table S5).

## Corpus-level reporting-completeness

The workflow returned scores for all 279 papers, with no failed or unscored cases. Mean completeness was 63.7/100 and median 67.5/100; 72 papers (25.8%) scored 50–64.9 and 130 (46.6%) scored 65–79.9 (Supplementary Tables S6 and S7).

Among applicable papers, incomplete reporting was most frequent for runtime/architecture (84.6%), governance/safety (82.8%), and provenance/reproducibility (78.1%). Evaluation and benchmark-process alignment were more complete, with incomplete rates of 25.8% and 29.8%, respectively (Table 2).

**Table 2. Corpus-level reporting completeness dimensions.**

| Dimension | Present, n | Partial, n | Missing, n | Not applicable, n | Incomplete among applicable, % |
|---|---|---|---|---|---|
| Runtime/architecture | 40 | 188 | 32 | 19 | 84.6 |
| Evaluation | 204 | 62 | 9 | 4 | 25.8 |
| Provenance/reproducibility | 61 | 189 | 28 | 1 | 78.1 |
| Governance/safety | 48 | 176 | 55 | 0 | 82.8 |
| Benchmark-process alignment | 191 | 68 | 13 | 7 | 29.8 |

***Incomplete among applicable*** *was calculated as the proportion of applicable papers labeled partial or missing. Papers labeled not applicable were excluded from the denominator.*

Thus, papers generally documented evaluation and benchmark procedures better than runtime architecture, reproducibility, and governance needed to interpret or reproduce the reported systems.

# Discussion

## Principal findings and implications

AGENT-O combines a modular ontology, semantic Agent Card profile, and reporting-completeness workflow. It represents runtime, models, workflow, tools, evaluation, provenance, governance, clinical-data use, and assessment records. Its central distinctions separate model role from specification and deployment, clinical use from manuscript statements, and assessment representation from the external workflow. This integrating layer adds agent-specific concepts without replacing established resources.

The corpus showed an evaluation–specification gap: evaluation and benchmark procedures were usually reported more completely than runtime/architecture, governance/safety, and provenance/reproducibility. Missing details included system configuration, human review, fallback, uncertainty, privacy, security, compliance, and traceability—information needed to interpret health-oriented agents responsibly.

AGENT-O therefore evaluates whether publications provide enough information to populate an ontology, not whether systems perform well or are clinically safe. This distinction supports author checklists, curation schemas, reviewer aids, repository metadata, and governance registries without conflating documentation with effectiveness.

## Limitations and future work

This study has three principal limitations. First, AGENT-O is a reusable framework rather than a complete or final ontology. Evolving agent architectures, clinical use cases, and governance requirements will require continued refinement. Its selective alignments support interoperability but do not replace or fully capture the semantics of the aligned resources.

Second, reporting-completeness scores assess publication content, not agent performance, clinical validity, fairness, safety, or deployment readiness. Third, the corpus was derived from two review-associated inventories rather than an independent systematic search, and its composition may affect the observed reporting-gap frequencies. The label-blinded GPT-5.1 workflow enabled source-grounded assessment at corpus scale, with source verification and deterministic checks reducing evidence errors, but labels were not adjudicated by multiple human reviewers. Results should therefore be interpreted as reproducible, ontology-guided, model-assisted estimates rather than a human-validated reference standard.

Future work will freeze and archive manuscript-associated releases, extend ontology terms and SHACL profiles when repeated gaps are observed, and quantify expert agreement and score error in a report-type- and score-stratified validation sample. We will also evaluate whether AGENT-O improves retrieval, structured annotation, and cross-study comparison.

# Conclusion

AGENT-O provides a reusable ontology, semantic Agent Card profile, and workflow for representing health-oriented AI agent systems and assessing publication-level reporting completeness. The corpus showed stronger reporting of evaluation and benchmark procedures than runtime architecture, governance, and provenance. By making system specifications, clinical use, evidence, and governance constraints queryable, AGENT-O supports structured annotation and cross-study comparison without implying agent performance or deployment readiness.

# Data and code availability

The AGENT-O public-release repository is available at https://github.com/Tao-AI-group/agent-o-health.

# Competing interests

The authors declare no competing interests.

# Funding

This work was supported by the National Institutes of Health under grant numbers **R01AG083039** and **R01AG084236**.